\documentclass[10pt, conference, compsocconf]{IEEEtran}
\ifCLASSINFOpdf
\else
\fi
\usepackage[utf8]{inputenc}
\usepackage[numbers,sort&compress,square]{natbib}
\usepackage{hyperref}
\usepackage{amsmath,amssymb,amsfonts}
\usepackage{lipsum,adjustbox}
\usepackage{algorithmic}
\usepackage{graphicx}
\usepackage{textcomp}
\usepackage{xcolor}
\usepackage{url}
\usepackage{multicol}
\usepackage{multirow}
\usepackage{hhline}
\usepackage[nomessages]{fp}

\usepackage{siunitx}

\usepackage{pgf-pie} 
\newcolumntype{C}[1]{>{\centering\let\newline\\\arraybackslash\hspace{0pt}}m{#1}}
\usepackage{pgfplots}
\newcommand{\footURL}[1]{\footnote{\url{#1}}}
\usepackage{tabularx,booktabs}
\newcolumntype{Y}{>{\centering\arraybackslash}X}
\usepackage{mathtools}
\usepackage{makecell}

\begin{document}
%
\title{Sentiment Analysis with Deep Learning Models: A Comparative Study on a Decade of Sinhala Language Facebook Data}


\author{\IEEEauthorblockN{Gihan Weeraprameshwara*, Vihanga Jayawickrama, Nisansa de Silva}
\IEEEauthorblockA{\textit{Department of Computer Science \& Engineering} \\
\textit{University of Moratuwa}\\
Moratuwa, Sri Lanka \\
* gihanravindu.17@cse.mrt.ac.lk}
\and
\IEEEauthorblockN{Yudhanjaya Wijeratne}
\IEEEauthorblockA{\textit{LIRNEasia}\\
Colombo, Sri Lanka}
}


%


\maketitle

\begin{abstract}
 The relationship between Facebook posts and the corresponding reaction feature is an interesting subject to explore and understand. To achieve this end, we test state-of-the-art Sinhala sentiment analysis models against a data set containing a decade worth of Sinhala posts with millions of reactions. For the purpose of establishing benchmarks and with the goal of identifying the best model for Sinhala sentiment analysis, we also test, on the same data set configuration, other deep learning models catered for sentiment analysis. In this study we report that the 3 layer Bidirectional LSTM model achieves an F1 score of 84.58\% for Sinhala sentiment analysis, surpassing the current state-of-the-art model; Capsule B, which only manages to get an F1 score of 82.04\%. Further, since all the deep learning models show F1 scores above 75\% we conclude that it is safe to claim that Facebook reactions are suitable to predict the sentiment of a text. 
\end{abstract}


%
\IEEEpeerreviewmaketitle

\section{Introduction}
\label{sec:introduction}

\newcommand{\re}{\textit{reaction sentiment data set} }

Understanding or analyzing human emotions could be quite a complex task for a computer. Nonetheless, the numerous possible applications of this task has driven researchers around the globe to attempt to tackle it through varying perspectives, causing sentiment analysis to become a leading research area in natural language processing. A wide range of fields including movie reviews~\cite{socher2013recursive}, product reviews~\cite{fang2015sentiment}, and political studies~\cite{rudkowsky2018more} benefit from advancements in this study area. 

However, a noteworthy gap in research on sentiment analysis, which is currently available, is the lack of interest in the perspective of the observer. Most research studies attempt to understand the sentiment expressed by the creator of the content. Only an extremely limited number of researchers such as~\citet{hui2017effects} have attempted to analyze the sentiment invoked by the content in an observer. While there are instances where the emotional reaction of the creator of content is more important than that of the observer, in certain instances such as political campaigns and brand management, understanding the reaction of the observer towards content could be quite important~\cite{jayawickrama2021seeking}.

Facebook reactions, which could be considered as user sentiment annotations for Facebook posts, contain an immense potential in the field of sentiment analysis from the perspective of the observer, due to the convenient availability of massive amounts of data.~\cite{pool2016distant,freeman2020measuring,graziani2019jointly,jayawickrama2021seeking}. 

Through this paper, we attempt to assess the applicability of Facebook data for sentiment analysis with respect to the Sinhala language. In contrast to English, for which numerous resources as well as previous research studies are readily available for sentiment analysis, the amount of research done in Sinhala in this arena remains highly unsatisfactory~\cite{de2019survey}. The fact that most of the research conducted on the subject are outdated, incomplete, or access-restricted contribute heavily to this issue~\cite{wijeratne2019natural}. To the best of our knowledge, the reigning state-of-the-art model for sentiment analysis in Sinhala is the \textit{Sencat} tool~\cite{demottesencatnodate}. The name of the tool is an abbreviation for \underline{Sen}timent \underline{Cat}egorization. The developers of the tool have tested the performance of several well-known sentiment analysis models for a data set consisting of over $15,059$ Sinhala news comments~\cite{senevirathne2020sentiment}. Using their results as a benchmark, we have tested the feasibility of Facebook data as a means for sentiment analysis in Sinhala. Here it should be noted that our \re is significantly (ten times) larger than that of~\citeauthor{senevirathne2020sentiment}. The \re contains $150,000$ Facebook posts with $235,660$ sentences as discussed in Section~\ref{sec:dataset}. 

Following the procedure adopted by \citet{senevirathne2020sentiment}, a comparison between different models: RNN~\cite{wang2016combination} , LSTM~\cite{hochreiter1997long}, GRU~\cite{chung2014empirical}, Bidirectional LSTM~\cite{schuster1997bidirectional}, combinations of baseline models and a convolutional neural network~\cite{wang2016combination}, 2 and 3 layer stacked LSTM~\cite{zhou2019sentiment}, 2 and 3 layer stacked Bidirectional LSTM~\cite{zhou2019sentiment}, HAHNN~\cite{abreu2019hierarchical}, capsule-A~\cite{zhao2018investigating}, and capsule-B~\cite{zhao2018investigating}, is conducted to select the model with the best performance. Additionally, two machine learning models introduced in the work of~\citet{jayawickrama2021seeking}; Core Reaction Set Model and Star Rating Model are also tested.

Results of this study indicate that the highest performance metrics are achieved by the 3-Layer Stacked BiLSTM model, with accuracy, recall, precision, and F1 score values of $83.59$, $83.59$, $85.59$, and $84.58$ respectively. These are significantly higher than the best results obtained by \citet{senevirathne2020sentiment}. Furthermore, in contrast to the results obtained by \citeauthor{senevirathne2020sentiment} where the best values of different performance metrics were scattered among different models, the consistent high performance of the 3-Layer Stacked BiLSTM model across all metrics provides further reassurance regarding the ability of the model to predict the sentiments included in the \re used in this study. The utilization of a larger data set is a major reason for this consistency as well as the significant increase in performance scores of all models tested.

\section{Background}
\label{Sec:Backgr}

Many of the currently available studies on sentiment analysis aim to understand the market for various products and services. 
The work of~\citet{fang2015sentiment} is seminal on product review analysis. It introduces a procedure for sentiment polarity categorization of product reviews, in which the sentiment analysis is carried out at a phrase-level. Phrases and words that convey a sentiment are identified as sentiment tokens, for which a sentiment score is calculated following the popular star rating model. Feature vectors developed using them are then used for predicting the sentiment polarity of previously unseen product reviews.
The study by~\citet{de2014sensing} also tackles the same problem, but at an aspect level. It calculates sentiment values separately for each aspect included in a review. An important property of this study is that it takes into account several semantic features such as negations, sentiment enhancements, sentiment shifts, and groups of words. This helps to increase the accuracy of transforming the sentiments embedded in text into mathematical values. The method proposed by~\citet{de2014sensing} has been tested for the corpus we derive the \re from by~\citet{jayawickrama2021seeking}.

\citet{wang2016combination} present a multitude of deep neural network architectures that combines CNNs and RNNs for sentiment analysis. These models exploit the key characteristics of both types of neural networks by using CNN layers to capture lower order features and RNN layers to capture higher order features. Though the study was originally conducted for capturing sentiments included in short text, the models included exhibit great performance in sentiment analysis even for longer text. 

This study was carried out using, \re which is a corpus composed of Facebook posts and their metadata. Numerous researchers have taken a similar path before, including \citet{pool2016distant} and \citet{freeman2019shared} who utilized Facebook data sets for sentiment detection. However, the work by \citet{freeman2019shared} is limited to a narrow scope of scholarly articles on Facebook. On the contrary, \citet{pool2016distant} has used a more generalized data set consisting of a wide range of sources. The authors explain the motive behind this choice as the need to choose the optimum sources to train machine learning models for each Facebook reaction. Additionally, the authors have conducted research on models with features such as TF-IDF, n-grams, and embeddings. A detailed comparison of the models is presented by them, which provides valuable insight into their use cases. 

While the aforementioned studies provide useful insight into sentiment analysis, there is a major barrier that our study has to face that many of them might find unfamiliar. Most of those research work relate to the English language, which is resource-rich. In contrast, our study relates to Sinhala language, which could be considered a resource-poor language in the NLP domain~\cite{wijeratne2019natural}. Research work related to sentiment analysis in Sinhala are highly limited, and most of the initiated research studies in this arena were abandoned or not released to the public~\cite{de2019survey}. 

The \textit{Sencat} sentiment analyzer by~\citet{demottesencatnodate} is the current state-of-the-art model for Sinhala sentiment analysis. The tool was originally trained using a data set comprising of $15,059$ purely Sinhala news comments. The sentiment classification of news comments is done at the document-level, under the assumption that the whole document represents a single emotion. 
\textit{Sencat} has been presented in two formats, namely:1) \textbf{Binary Class Classification:} \textit{Positive} and \textit{Negative} 2) \textbf{Multi Class Classification:} \textit{Positive}, \textit{Negative}, \textit{Neutral}, and \textit{Conflict}
In the case of multi class classification, news comments that display neither positive nor negative emotion are assigned to the \textit{Neutral} class while comments that display both positive and negative emotions are classified as \textit{Conflict}.
Prior to introducing sentiment categorization,~\citet{demottesencatnodate} have tested the performance of several sentiment analysis tools built using deep learning techniques against a data set composed of Sinhala news comments~\cite{senevirathne2020sentiment} which they published earlier. 

\citet{jayawickrama2021seeking}, on the other hand, have introduced a set of simple machine learning models for sentiment analysis in Sinhala.In which they have introduced three models that have been developed specifically targeting sentiment analysis with Facebook data, and has been tested with the \re which we use in this study. A simple procedure of using the average of sentiment vectors of previously seen data to predict that of unseen data is utilized in these models. \textit{All Reaction Set Model} and \textit{Core Reaction Set Model} attempt to predict the distribution of Facebook reactions while the \textit{Star Rating Model} concatenates reactions into a single value representing where it would fall in the scale between extreme positivity and extreme negativity.
%
The work of \citet{medagoda2015sentiment}, although dated a few years back, provides an insightful exploration of the applicability of tools available for resource-rich languages to resource-poor languages such as Sinhala. 

\section{Methodology}
\label{Sec:Meth}

In this section we discuss how we create \re using the Facebook data set composed by \citet{wijeratne2020sinhala} and test it on the sentiment analysis models proposed by \citet{senevirathne2020sentiment} and \citet{jayawickrama2021seeking} where the former includes the model that obtains state-of-the-art results for direct Sinhala sentiment analysis. 


\subsection{Data Set}
\label{sec:dataset}

This study is conducted using a portion of the Facebook data set composed by~\citet{wijeratne2020sinhala}. It was selected as the most suitable for the study for a multitude of reasons. Firstly, Facebook, unlike most other social medial platforms, enables the users to annotate posts with a set of sentimental reactions that are representative of both positive and negative sentiments. At the time the data set was composed, a total of 7 Facebook reactions were available: \textit{Like}, \textit{Love}, \textit{Wow}, \textit{Haha}, \textit{Thankful}, \textit{Sad}, and \textit{Angry}. These reactions chosen to be applied to posts by Facebook users can be considered as a user sentiment annotation for posts, eliminating the need for a separate dedicated manual sentiment annotation of data sets, which is a tedious and highly resource consuming task. These reactions have a direct correlation to the actual sentiment invoked in a user by the post, which means, the reaction given by a user can be taken as the sentiment that particular user was expressing with a high confidence as opposed to a third party guessing and tagging a data set after the fact~\cite{tian-etal-2017-facebook}. However, due to behaviours such as sarcastic reactions~\cite{jayawickrama2021seeking}, it should be noted that Facebook reactions may not have a straightforward one-to-one mapping to the actual sentiment experienced by the users 100\% of the times. Nevertheless, it is still a more accurate measure of reaction sentiment than what a third party may tag. The deviant reactions, such as  sarcastic reactions, are an interesting aspect to explore in future studies.

The original data set created by~\citet{wijeratne2020sinhala} consists of $1,820,930$ Facebook posts taken from pages popular in Sri Lanka over a time period ranging from 01-01-2010 to 02-02-2020. Post text as well as several metadata, including reaction counts for each of the Facebook reactions, are among the features present in the data set. 
For this study, a subset of $150,000$ posts, which we refer to as \re has been extracted from the original corpus. The counts and the percentages of each reaction included in this reduced data set is presented in Table~\ref{Table:Counts}.

\FPeval{\Like}{clip(38889706)}
\FPeval{\Love}{clip(2336796)}
\FPeval{\Wow}{clip(321178)}
\FPeval{\Haha}{clip(1486413)}
\FPeval{\Sad}{clip(609597)}
\FPeval{\Angry}{clip(349296)}
\FPeval{\Thankful}{clip(7)}

\FPeval{\Total}{clip(\Like+\Love+\Wow+\Haha+\Sad+\Angry+\Thankful)}

\FPeval{\Likes}{clip(round((100*\Like)/\Total,3))}
\FPeval{\Loves}{clip(round((100*\Love)/\Total,2))}
\FPeval{\Wows}{clip(round((100*\Wow)/\Total,2))}
\FPeval{\Hahas}{clip(round((100*\Haha)/\Total,2))}
\FPeval{\Sads}{clip(round((100*\Sad)/\Total,2))}
\FPeval{\Angrys}{clip(round((100*\Angry)/\Total,2))}
\FPeval{\Thankfuls}{clip(round((100*\Thankful)/\Total,3))}

\FPeval{\TotalConsidered}{clip(\Love+\Wow+\Sad+\Angry)}

\FPeval{\PortionLove}{clip(round((100*\Love)/\TotalConsidered,2))}
\FPeval{\PortionWow}{clip(round((100*\Wow)/\TotalConsidered,2))}
\FPeval{\PortionHaha}{clip(round((100*\Haha)/\TotalConsidered,2))}
\FPeval{\PortionSad}{clip(round((100*\Sad)/\TotalConsidered,2))}
\FPeval{\PortionAngry}{clip(round((100*\Angry)/\TotalConsidered,2))}


\begin{table}[!htb]
\renewcommand{\arraystretch}{1.3}
\centering
\caption{Total counts of reactions in the selected portion of the data set.}
\label{Table:Counts}
\begin{tabular}{l r r r}
\hline
 \multirow{2}{*}{\textbf{Reaction}} & \multirow{2}{*}{\textbf{Count}} & \multicolumn{2}{c}{\textbf{Percentage}} \\
 \hhline{~~--}
 & & \textbf{Original} & \textbf{Filtered} \\
 \hline
 Like & \Like & \Likes& - \\
 Love & \Love& \Loves & \PortionLove\\
 Wow & \Wow & \Wows  & \PortionWow \\
 Haha & \Haha & \Hahas & \PortionHaha \\
 Sad & \Sad & \Sads & \PortionSad \\
 Angry & \Angry & \Angrys & \PortionAngry \\
 Thankful & \Thankful& \Thankfuls & - \\
 \hline
\end{tabular}
\end{table}

\subsection{Pre-processing}
\label{sec:prepro}

The version of the data set that \citet{wijeratne2020sinhala} had made available has been pre-processed by them to a certain extent. However, we observed that their procedure has resulted in some characters of the Sinhala language being removed as well. Therefore, we obtained the raw corpus and performed pre-processing procedures better tailored per our requirements.

The corpus used needed several pre-processing steps in order to conform to the requirements of this study. The \textit{Message} field contained raw, unprocessed text including non-printable characters, special characters, and text in various languages that would not be useful within the scope.

\begin{figure}
\centering
\includegraphics[width=0.45\textwidth]{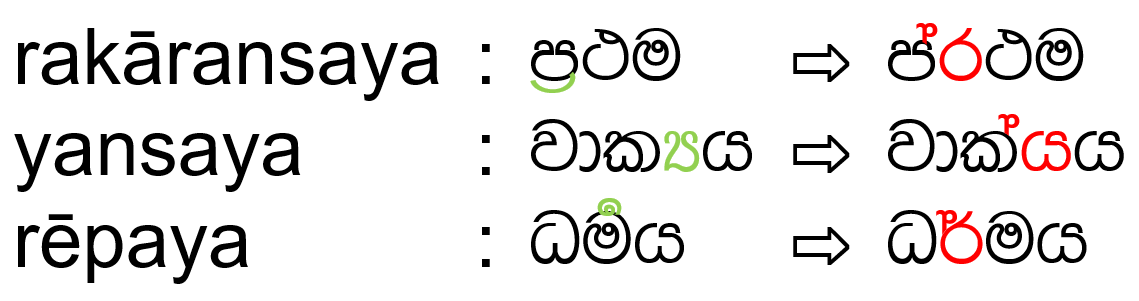}
    \caption{Examples of \textit{rakāransaya}, \textit{yansaya}, and \textit{rēpaya} being replaced by other unicode characters}
    \label{fig:unicode}
\end{figure}

Non printable characters belonging to the unicode categories \textit{Cc}, \textit{Cn}, \textit{Co}, and \textit{Cs} were substituted with white spaces~\cite{davis2008unicode}. The \textit{Cf} category too was replaced in the same manner, with the exception of the character \textit{Zero Width Joiner}, which often appears in the middle of Sinhala words consisting of characters such as \textit{rakāransaya}, \textit{yansaya}, and \textit{rēpaya} shown in the Figure~\ref{fig:unicode} . 
Singular unicode characters to represent these characters have not been introduced due to them being derivatives of existing Sinhala unicode characters. Thus a method of including the \textit{Zero Width Joiner} between characters that phonetically represent the said characters is often utilized in Sinhala text to indicate their occurrence. 
As such, replacing the \textit{Zero Width Joiner} with a space could lead to erroneous tokenization of Sinhala words. Therefore, it was replaced by a null string instead.

From the data set thus filtered, email addresses, URLs, user-tags (format: \texttt{@user}), and hashtags were removed. Numerical content too was filtered out as they do not possess a significant sentimental value. Since the corpus used by \citet{senevirathne2020sentiment} was composed exclusively of Sinhala text, to facilitate a fair comparison, any words that included characters which did not belong to the Sinhala language were removed from the corpus. Furthermore, the set of stop words for the Facebook data set generated by \citet{wijeratne2020sinhala} were removed as well, since stop words possess a rather low information value in the context of sentiment analysis~\cite{luhn1957statistical}. As the final pre-processing step, multiple continuous white spaces were replaced by a single white space.

\subsection{Normalization and Annotation}
\label{sec:anno}

The set of deep learning models used by \citet{senevirathne2020sentiment} perform a binary classification of text, classifying them as either \textit{Positive} or \textit{Negative} based on a sentimental perspective. To evaluate the performance of our corpus against those models, the Facebook reactions needed to be integrated into a single binary value. To achieve this end, only the \textit{Love}, \textit{Wow}, \textit{Sad}, and \textit{Angry} reactions were considered, omitting the \textit{Like}, \textit{Thankful}, and \textit{Haha} reactions. 

\textit{Like} and \textit{Thankful} reactions could be considered as outliers in comparison to the other reactions. In the reduced data set used for this study, a total count of $38,889,706$ \textit{Like} reactions were present, which amounted to $88.4\%$ of the total reaction count. In contrast, only $7$ \textit{Thankful} reactions were present in the corpus, which is a negligible value in comparison to the counts of other reactions. Inclusion of these reactions in the target label annotation would introduce a massive imbalance to the data set. The unusual behaviour of these two reactions could be attributed to the duration they were present on the Facebook platform~\cite{jayawickrama2021seeking}. The percentages of the remaining reactions after removing \textit{Like} and \textit{Thankful} is shown under the \textit{Filtered} column in Table~\ref{Table:Counts}.

Furthermore, the sentiment depicted by the \textit{Haha} reaction is a controversial matter as Facebook users tend to use it both genuinely and ironically, causing the sentiment category of the reaction to vary between \textit{Positive} and \textit{Negative} though it was originally introduced to Facebook as a positive reaction~\cite{jayawickrama2021seeking}. Thus, the \textit{Haha} reaction too has been excluded in the binary classification of reactions.
\textit{Love} and \textit{Wow} are considered as \textit{Positive} while \textit{Sad} and \textit{Angry} are considered as \textit{Negative}. Figure~\ref{fig:PreprocessPie} portrays the distribution of the considered reactions in the data set.

\newcommand{\StarRadius}{2}

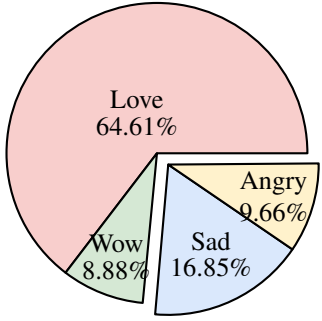
\begin{figure}
    \centering
\definecolor{wedgeLove}{RGB}{ 248  206  204}
\definecolor{wedgeWow}{RGB}{ 213  232  212}
\definecolor{wedgeSad}{RGB}{ 218  232  252}
\definecolor{wedgeAngry}{RGB}{ 255  242  204}
    
    \begin{tikzpicture}
    \pie[color = {wedgeLove,wedgeWow}, 
        radius=\StarRadius,
        text = inside
        ]
        {\PortionLove/Love,
        \PortionWow/Wow}
    \pie[pos={0.15,-0.15},
        color = {wedgeSad,wedgeAngry},
        rotate = 265,
        radius=\StarRadius,
        text = inside,
        sum = 100]
        {\PortionSad/Sad,
        \PortionAngry/Angry}
    \end{tikzpicture}
    \caption{Reaction Percentages after filtering out \textit{Like}, \textit{Thankful} and \textit{Haha} reactions.}
    \label{fig:PreprocessPie}
\end{figure}

Since the focus of this study is to analyze the distribution of reactions received by a post, inclusion of the raw number of reactions could create an unwelcome bias towards posts with high reaction counts. To overcome this issue, steps were taken to normalize the reaction counts of posts. For a given Facebook post, the summation of reactions $t$ is taken using the count of reactions \textit{Love}($c_l$), \textit{Wow}($c_w$), \textit{Sad}($c_s$), and \textit{Angry}($c_a$) as demonstrated through Equation~\ref{Eq:tot}. The normalized value $n_r$ of a given reaction $r$ for a certain Facebook post is obtained by dividing the reaction count of the said reaction $c_r$ by the total reaction count $t$ of the post.  

\begin{equation}
\label{Eq:tot}
    t=c_l+c_w+c_s+c_a
\end{equation}

\begin{equation}
\label{Eq:norm}
    n_r=\frac{c_r}{t}
\end{equation}

The normalized value of positive reactions for a post is obtained using the \textit{Love} and \textit{Wow} reactions, as shown by Equation~\ref{Eq:pos}. The summation of normalized values of the aforementioned reactions are assigned as the positive value $pos$ of a post. 
Similarly, the negative value $neg$ of a Facebook post is obtained by the summation of normalized \textit{Sad} and \textit{Angry} reaction values. This computation is shown in Equation~\ref{Eq:neg}.

\begin{equation}
\label{Eq:pos}
    pos=n_l + n_w
\end{equation}

\begin{equation}
\label{Eq:neg}
    neg=n_s + n_a
\end{equation}

The net sentiment value $sen$ of a Facebook post is then calculated by subtracting the negative value $neg$ of the post from the positive value \textit{pos}, as shown in Equation~\ref{Eq:sen} where $-1\leq sen\leq 1$ holds due to the normalization done by Equation~\ref{Eq:norm}.

\begin{equation}
\label{Eq:sen}
    sen={pos}-{neg}
\end{equation}

This sentiment value is then utilized to determine the sentiment category to which the post is assigned. Equation~\ref{Eq:label} shows the classification process.

\begin{equation}
\label{Eq:label}
\text{Label =}\begin{cases}
    Positive & \text{if } sen \geq 0 \\
    Negative & \text{if } sen < 0 \\
    \end{cases}
\end{equation}


It should be noted that~\citet{senevirathne2020sentiment} presents two classification methodologies: binary and multi-class. However, the multi-class classification does not befit our study given that their definitions of categories do not match the nature of our data set. They define the \textit{Neutral} class as text that possess neither positive nor negative sentiment, and the \textit{Conflict} class as text that possess both positive and negative sentiment. Only text that exclusively exhibit positive or negative emotions are classified as \textit{Positive} or \textit{Negative} in this classification. 
In our corpus, Facebook posts that have neither positive nor negative reactions would be posts that have not received any reactions from the considered classes, which depicts more of a lack of reaction data than a true sense of neutrality. Furthermore, a large proportion of posts would be allocated into the \textit{Conflict} class as the same post that invokes a positive reaction in one Facebook user often has the capability to invoke a negative reaction in another user, and vice versa. An argument may be raised here to say the instances where \textit{sen} value of Equation~\ref{Eq:sen} equal to $0$ might map to the union of \textit{Neutral} and \textit{Conflict} classes. However that definition is extremely narrow given that the probability of a post having exactly equal $pos$ and $neg$ values are extremely low. Thus, if the $sen=0$ rule is used, we will be introducing an extreme minority class. As we saw on the discussion on the \textit{Thankful} reaction in~\citet{jayawickrama2021seeking} as well as Section~\ref{sec:anno}, this would have resulted in an undesirable outcome.

Thus, the binary classification where text that exhibits more positive sentiment than negative is classified as \textit{Positive} and vice versa is selected for this study.
Upon completion of those steps, $116,061$ of the $150,000$ Facebook posts included in the data set were classified as \textit{Positive} and the remaining $33,939$ were classified as \textit{Negative}.

\subsection{Model Training}
\label{sec:train}

Following the conventions set by the benchmark~\citet{senevirathne2020sentiment}, we next use the fastText~\cite{bojanowski2017enriching,joulin2016bag} model with 300 dimensions trained using the aforementioned Sinhala News Comments data set to convert the post text included in each Facebook post to vectors.
%
The prepared data set is then used to train each deep learning model tested by \citet{senevirathne2020sentiment} and two models introduced by \citet{jayawickrama2021seeking}; \textit{Core Reaction Set Model} and \textit{Star Rating Model}. 


\FPeval{\ORNNA}{58.98}
\FPeval{\ORNNR}{54.98}
\FPeval{\ORNNP}{42.93}
\FPeval{\ORNNF}{42.30}

\FPeval{\OLSTMA}{62.88}
\FPeval{\OLSTMR}{51.93}
\FPeval{\OLSTMP}{70.95}
\FPeval{\OLSTMF}{54.50}

\FPeval{\OGRUA}{62.78}
\FPeval{\OGRUR}{62.78}
\FPeval{\OGRUP}{60.93}
\FPeval{\OGRUF}{54.83}

\FPeval{\OBiLSTMA}{63.81}
\FPeval{\OBiLSTMR}{63.81}
\FPeval{\OBiLSTMP}{61.17}
\FPeval{\OBiLSTMF}{57.71}

\FPeval{\OCNNGRUA}{61.59}
\FPeval{\OCNNGRUR}{61.59}
\FPeval{\OCNNGRUP}{60.41}
\FPeval{\OCNNGRUF}{54.19}

\FPeval{\OCNNLSTMA}{61.89}
\FPeval{\OCNNLSTMR}{61.89}
\FPeval{\OCNNLSTMP}{57.82}
\FPeval{\OCNNLSTMF}{55.30}

\FPeval{\OCNNBiLSTMA}{62.72}
\FPeval{\OCNNBiLSTMR}{62.72}
\FPeval{\OCNNBiLSTMP}{59.54}
\FPeval{\OCNNBiLSTMF}{58.53}

\FPeval{\OLSTMTWOA}{61.92}
\FPeval{\OLSTMTWOR}{61.92}
\FPeval{\OLSTMTWOP}{56.92}
\FPeval{\OLSTMTWOF}{53.17}

\FPeval{\OLSTMTHREEA}{62.48}
\FPeval{\OLSTMTHREER}{62.48}
\FPeval{\OLSTMTHREEP}{54.76}
\FPeval{\OLSTMTHREEF}{53.67}

\FPeval{\OBiLSTMTWOA}{63.18}
\FPeval{\OBiLSTMTWOR}{63.18}
\FPeval{\OBiLSTMTWOP}{60.50}
\FPeval{\OBiLSTMTWOF}{57.78}

\FPeval{\OBiLSTMTHREEA}{63.13}
\FPeval{\OBiLSTMTHREER}{46.63}
\FPeval{\OBiLSTMTHREEP}{69.71}
\FPeval{\OBiLSTMTHREEF}{59.42}

\FPeval{\OHAHNNA}{61.16}
\FPeval{\OHAHNNR}{48.54}
\FPeval{\OHAHNNP}{71.08}
\FPeval{\OHAHNNF}{59.25}

\FPeval{\OCapAA}{61.89}
\FPeval{\OCapAR}{61.89}
\FPeval{\OCapAP}{56.12}
\FPeval{\OCapAF}{53.55}

\FPeval{\OCapBA}{63.23}
\FPeval{\OCapBR}{63.23}
\FPeval{\OCapBP}{59.84}
\FPeval{\OCapBF}{59.11}


\FPeval{\RNNA}{77.90}
\FPeval{\RNNR}{77.90}
\FPeval{\RNNP}{77.54}
\FPeval{\RNNF}{77.72}

\FPeval{\LSTMA}{81.53}
\FPeval{\LSTMR}{81.53}
\FPeval{\LSTMP}{80.95}
\FPeval{\LSTMF}{81.24}

\FPeval{\GRUA}{81.28}
\FPeval{\GRUR}{81.28}
\FPeval{\GRUP}{81.39}
\FPeval{\GRUF}{81.33}

\FPeval{\BiLSTMA}{82.58}
\FPeval{\BiLSTMR}{82.58}
\FPeval{\BiLSTMP}{82.58}
\FPeval{\BiLSTMF}{82.58}

\FPeval{\CNNGRUA}{81.17}
\FPeval{\CNNGRUR}{81.17}
\FPeval{\CNNGRUP}{81.57}
\FPeval{\CNNGRUF}{81.37}

\FPeval{\CNNLSTMA}{81.47}
\FPeval{\CNNLSTMR}{81.47}
\FPeval{\CNNLSTMP}{82.07}
\FPeval{\CNNLSTMF}{81.78}

\FPeval{\CNNBiLSTMA}{80.98}
\FPeval{\CNNBiLSTMR}{80.98}
\FPeval{\CNNBiLSTMP}{81.01}
\FPeval{\CNNBiLSTMF}{81.00}

\FPeval{\LSTMTWOA}{81.87}
\FPeval{\LSTMTWOR}{81.87}
\FPeval{\LSTMTWOP}{81.29}
\FPeval{\LSTMTWOF}{81.58}

\FPeval{\LSTMTHREEA}{81.38}
\FPeval{\LSTMTHREER}{81.39}
\FPeval{\LSTMTHREEP}{81.09}
\FPeval{\LSTMTHREEF}{81.24}

\FPeval{\BiLSTMTWOA}{82.49}
\FPeval{\BiLSTMTWOR}{82.49}
\FPeval{\BiLSTMTWOP}{82.63}
\FPeval{\BiLSTMTWOF}{82.56}

\FPeval{\BiLSTMTHREEA}{83.59}
\FPeval{\BiLSTMTHREER}{83.59}
\FPeval{\BiLSTMTHREEP}{85.59}
\FPeval{\BiLSTMTHREEF}{84.58}

\FPeval{\HAHNNA}{77.27}
\FPeval{\HAHNNR}{77.39}
\FPeval{\HAHNNP}{77.39}
\FPeval{\HAHNNF}{77.39}

\FPeval{\CapAA}{80.44}
\FPeval{\CapAR}{80.44}
\FPeval{\CapAP}{78.91}
\FPeval{\CapAF}{79.67}

\FPeval{\CapBA}{82.51}
\FPeval{\CapBR}{82.51}
\FPeval{\CapBP}{81.58}
\FPeval{\CapBF}{82.04}

\FPeval{\CoreA}{52.45}
\FPeval{\CoreR}{85.13}
\FPeval{\CoreP}{35.19}
\FPeval{\CoreF}{49.80}

\FPeval{\StarA}{69.17}
\FPeval{\StarR}{22.36}
\FPeval{\StarP}{68.96}
\FPeval{\StarF}{33.77}

\begin{table*}[!htb]
\renewcommand{\arraystretch}{1.3}
\centering
\caption{Results of each model}
\label{Table:ModelResult}
\begin{tabularx}{\textwidth}{l *{3}{Y}Y*{4}{Y} }
\hline
 \multirow{2}{*}{\textbf{Model}} &
 \multicolumn{4}{c}{\makecell{\textbf{\citet{senevirathne2020sentiment} (\%)} \\ \textbf{(10 fold cross validation)}}} & \multicolumn{4}{c}{\makecell{\textbf{This Study (\%)} \\ \textbf{(Holdout method)}}} \\
 \hhline{~--------}
 & \textbf{A} & \textbf{R} & \textbf{P} & \textbf{F1} & \textbf{A} & \textbf{R} & \textbf{P} & \textbf{F1}\\
 \hline
Core Reaction Model~\cite{jayawickrama2021seeking} & - & - & - & - & \CoreA & \CoreR & \CoreP & \CoreF  \\
 Star Rating Model~\cite{de2014sensing,jayawickrama2021seeking} & - & - & - & -  & \StarA  & \StarR  & \StarP  & \StarF  \\
 \hline
 RNN~\cite{wang2016combination}   & \ORNNA & \ORNNR & \ORNNP & \ORNNF & \RNNA & \RNNR & \RNNP & \RNNF  \\
 GRU~\cite{chung2014empirical}   & \OGRUA & \OGRUR & \OGRUP & \OGRUF & \GRUA & \GRUR & \GRUP & \GRUF  \\
 LSTM~\cite{hochreiter1997long}  & \OLSTMA & \OLSTMR & \OLSTMP & \OLSTMF & \LSTMA & \LSTMR & \LSTMP & \LSTMF  \\
 BiLSTM~\cite{schuster1997bidirectional}   & \textbf{\OBiLSTMA} & \textbf{\OBiLSTMR} & \OBiLSTMP & \OBiLSTMF & \BiLSTMA & \BiLSTMR & \BiLSTMP & \BiLSTMF\\
 \hline
 CNN~\cite{wang2016combination} + GRU~\cite{chung2014empirical} & \OCNNGRUA & \OCNNGRUR & \OCNNGRUP & \OCNNGRUF & \CNNGRUA & \CNNGRUR & \CNNGRUP & \CNNGRUF  \\
 CNN~\cite{wang2016combination} + LSTM~\cite{hochreiter1997long}   & \OCNNLSTMA & \OCNNLSTMR & \OCNNLSTMP & \OCNNLSTMF & \CNNLSTMA & \CNNLSTMR & \CNNLSTMP & \CNNLSTMF  \\
 CNN~\cite{wang2016combination} + BiLSTM~\cite{schuster1997bidirectional}  & \OCNNBiLSTMA & \OCNNBiLSTMR & \OCNNBiLSTMP & \OCNNBiLSTMF & \CNNBiLSTMA & \CNNBiLSTMR & \CNNBiLSTMP & \CNNBiLSTMF  \\
  \hline
 Stacked LSTM 2~\cite{zhou2019sentiment}  & \OLSTMTWOA & \OLSTMTWOR  & \OLSTMTWOP & \OLSTMTWOF & \LSTMTWOA & \LSTMTWOR & \LSTMTWOP & \LSTMTWOF \\
 Stacked LSTM 3~\cite{zhou2019sentiment}  & \OLSTMTHREEA & \OLSTMTHREER & \OLSTMTHREEP & \OLSTMTHREEF & \LSTMTHREEA & \LSTMTHREER & \LSTMTHREEP & \LSTMTHREEF \\
 Stacked BiLSTM 2~\cite{zhou2019sentiment}  & \OBiLSTMTWOA & \OBiLSTMTWOR & \OBiLSTMTWOP & \OBiLSTMTWOF & \BiLSTMTWOA & \BiLSTMTWOR & \BiLSTMTWOP & \BiLSTMTWOF \\
 Stacked BiLSTM 3~\cite{zhou2019sentiment}  & \OBiLSTMTHREEA & \OBiLSTMTHREER & \OBiLSTMTHREEP & \textbf{\OBiLSTMTHREEF} & \textbf{\BiLSTMTHREEA} & \textbf{\BiLSTMTHREER} & \textbf{\BiLSTMTHREEP} & \textbf{\BiLSTMTHREEF} \\
 \hline
 HAHNN~\cite{abreu2019hierarchical}  & \OHAHNNA & \OHAHNNR & \textbf{\OHAHNNP} & \OHAHNNF & \HAHNNA & \HAHNNR & \HAHNNP & \HAHNNF  \\
 \hline
 Capsule-A~\cite{zhao2018investigating} & \OCapAA & \OCapAR & \OCapAP & \OCapAF & \CapAA & \CapAR & \CapAP & \CapAF  \\
 Capsule-B~\cite{zhao2018investigating}  & \OCapBA & \OCapBR & \OCapBP & \OCapBF & \CapBA & \CapBR & \CapBP & \CapBF  \\
 \hline
 
\end{tabularx}
\end{table*}

\section{Results}
\label{Sec:Res}

The results obtained from this study following the steps described in Section~\ref{Sec:Meth} is presented in Table~\ref{Table:ModelResult} along with a summarized view of the results reported by \citet{senevirathne2020sentiment} for the purpose of comparison. Here, the performance measures accuracy, recall, precision, and F1 score of each model are displayed in the columns \textit{A},\textit{R},\textit{P}, and \textit{F1} respectively. 

Upon comparing the results of the two research studies, a clear increment in all performance metrics could be observed in the results of this study. The highest F1 score reported by \citeauthor{senevirathne2020sentiment} is $59.42$, while the lowest F1 score received by a deep learning model in this study is $77.39$. Only the simple models by \citeauthor{jayawickrama2021seeking} exhibit F1 scores lower than the peak F1 score of \citeauthor{senevirathne2020sentiment}. It should be noted that the performance of the Core Reaction Model, though lagging behind in comparison to the results obtained through this study, lies in a comparable range with the deep learning models in the results of \citeauthor{senevirathne2020sentiment}, with a F1 score of $49.80$. The highest F1 score observed in this study reaches as high as $84.58$.

In contrast to the results reported by \citet{senevirathne2020sentiment}, the 3 Layer BiLSTM model exhibits the best results for every performance metric in this study, with accuracy, recall, precision, and F1 score values of $83.59\%$, $83.59\%$, $85.59\%$, and $84.58\%$ respectively. The BiLSTM model which produced the best accuracy and recall values in \citet{senevirathne2020sentiment} has proved its worth with accuracy and recall values of $82.58\%$ each, which are the second highest values observed in those metrics. However, the HAHNN model which exhibited the highest precision score in \citet{senevirathne2020sentiment} has slumped down to the 13th place with a precision value of $77.54\%$. This model displays the lowest increase in performance metrics upon comparing the results of the two studies. The 3 Layer Stacked BiLSTM model, which has produced the highest F1 score in the work of \citeauthor{senevirathne2020sentiment}, has continued to excel achieving the highest values in all performance metrics in this study. The state-of-the-art model; Capsule-B highlighted by \citeauthor{senevirathne2020sentiment} as the best in the set, has been pushed down to the 4th rank with respect to the F1 score in the new results, outperformed by the BiLSTM model as well as the 2 and 3 Layer Stacked BiLSTM models. It is noteworthy that all BiLSTM models except for the CNN + BiLSTM model has achieved remarkable F1 scores in our experiments.

The performance of the baseline models; GRU, LSTM, and BiLSTM have experienced a drop when a CNN layer is added to the model in both experiments. This counter-intuitive result may be a result of the specific configuration of CNN that \citet{senevirathne2020sentiment} proposed. However, given that we needed to do a fair comparison with the results they have reported, when used for \re we did not vary the CNN in this study. We discuss this further in Section~\ref{Sec:Con}. 
The lowest performance across both studies is observed in the Star Rating Model, with a F1 score of $33.77\%$. Though the model has achieved a fairly decent precision value of $68.96\%$, the rather weak recall value of $22.36\%$ has pushed the model down to this position.

It should be highlighted is that the 10-fold cross validation method was used to obtain the performance measures reported by \citeauthor{senevirathne2020sentiment}. Since this study uses $150,000$ rows from the Facebook data set in contrast to the $15,059$ rows used in the work of~\citeauthor{senevirathne2020sentiment}, the excessive resource consumption of using 10-fold cross validation was too costly. Therefore, the holdout method was used with the data set split into development and test sets in the ratio 8:2, with the development set again divided into train and validation sets in the ratio 9:1. However, even with this setback to training, this study achieved better results than~\citet{senevirathne2020sentiment} as shown in Table~\ref{Table:ModelResult}.

\section{Conclusion and Future Work}
\label{Sec:Con}

\pgfplotsset{compat=1.14}

\begin{figure*}[!htb]
\centering
\begin{tikzpicture}
\pgfplotsset{
width=0.95\textwidth,
height=0.3\textwidth,
every axis legend/.append style={
at={(0.5,1.03)},
anchor=south,
},
}
\begin{axis}[
    legend columns=3,
    axis lines=middle,
    xmin=0,
    xmax=17,
    ymin=20,
    x label style={
        at={(axis description cs:0.5,-0.6)},
        below=7mm},
    y label style={
        at={(axis description cs:-0.10,0.5)},
        rotate=90,
        anchor=south},
    legend style={above=2mm},
    xlabel=Models,
    ylabel=F1 score (\%) ,
    enlargelimits = false,
    xticklabels from table={F1Plot.dat}{Model},
    xtick=data,
    xticklabel style = {rotate=90,anchor=east}]
\addplot[cyan!60!green!80!blue,thick,mark=square*] table [y={This Study},x=X]{F1Plot.dat};
\addlegendentry{This Study}
\addplot[pink!60!purple,thick,mark=square*] table [y={Senevirathne et al.},x=X]{F1Plot.dat};
\addlegendentry{Senevirathne et al.}
\end{axis}
\end{tikzpicture}
\caption{Change of the F1 score of the Models.}
\label{fig:F1plot}
\end{figure*}
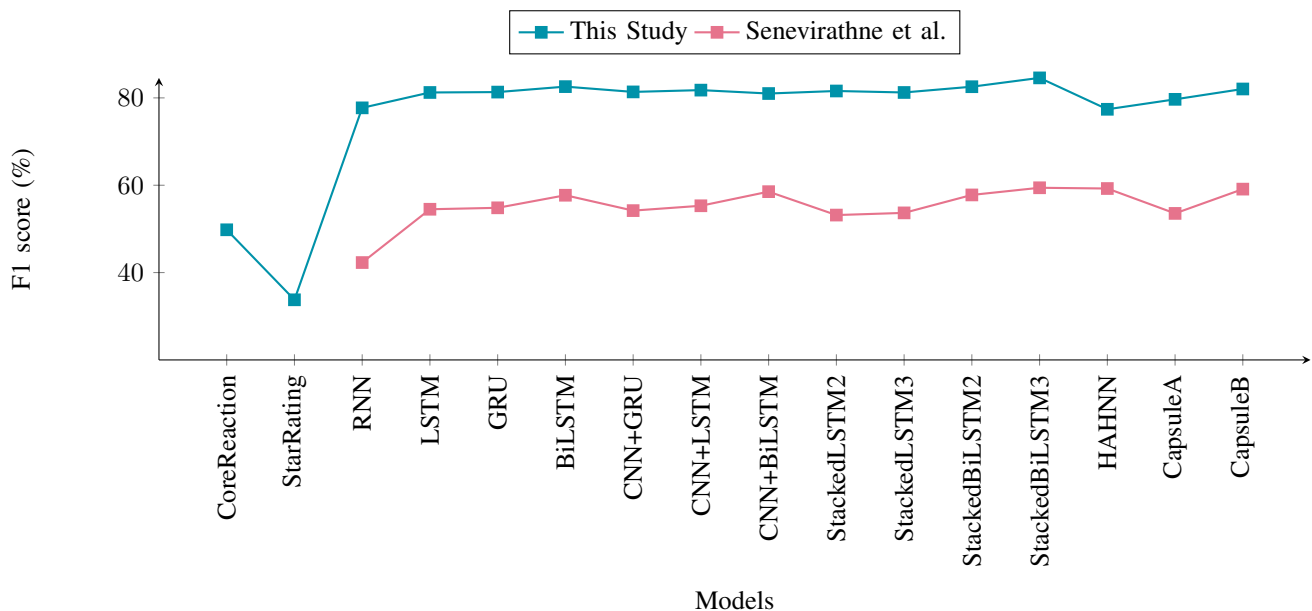
The results of this study exhibit a clear improvement of performance metrics in comparison to the work done by \citet{senevirathne2020sentiment}. The differences among F1 scores of the two studies are visualized in Fig.~\ref{fig:F1plot}. The size of the selected data set can be postulated as a major reason for the drastic improvement of results in this study; the data set we used is almost 10 times the size of the corpus used by \citet{senevirathne2020sentiment}. As deep learning is generally data hungry~\cite{marcus2018deep}, the $15,059$ entries in their data set may have been insufficient to properly train the models.
The fact that a larger user annotated data set could produce better performance than a smaller manually annotated data set is a promising result for resource poor languages, for which manually annotated data sets are rather difficult to create or find in comparison to user annotated social media data sets. Even in the cases where manually annotated data is available, in configurations such as parallel corpora, it has been observed that the data is of poor quality for Sinhala~\cite{caswell2021quality}.  

The BiLSTM models outshine the rest of the models in the experiment, with the exception of the BiLSTM model with a CNN layer. This effect could be due to the ability of BiLSTM models to be trained in both positive and negative time directions~\cite{schuster1997bidirectional}
, which is countered by upper layers in the stacked models filtering out contextual information from positive and negative time sequences when the model is paired with a CNN~\cite{zhou2019sentiment}.

In the work of~\citet{senevirathne2020sentiment}, the reason for the lack of performance of baseline models paired with a CNN layer is explained as the lack of data entries to train the model properly. Furthermore, since only one layer of one dimensional CNN layer is used, a different configuration of the CNN layer may alter the results. However, since changing the CNN layer configuration may lead to unfair comparison with the model used in the work of~\citeauthor{senevirathne2020sentiment} no adjustment was done as discussed earlier in Section~\ref{Sec:Res}.
The HAHNN model~\cite{abreu2019hierarchical} shows the lowest improvement in the performance compared to other models. Since the HAHNN model focuses on the attention mechanism introduced by the work of~\citet{vaswani2017attention}, the lack of respect to strict grammatical structures observed in colloquial text in our corpus may have hindered the capability of HAHNN.  

The Core Reaction Set model and the Star Rating model taken from~\citet{jayawickrama2021seeking}, being derived from classical statistical modeling, lack the complexity to perform well in the sentiment analysis task compared to other models. Thus, further improvements will be required in order for them to become effective sentiment analysis models. 

The word embedding model used in this work has been trained using the Sinhala News Comments data set. This can be replaced by a new word embedding trained using the Facebook data set itself, which could capture the relationships among words in the data set more effectively. Different word embedding models such as fastText, Word2vec~\cite{mikolov2013efficient}, and Glove~\cite{pennington2014glove} with different dimensions could be tested to select the best embedding. 

The latest model known as \textit{transformers} introduced by \citet{vaswani2017attention} is another area to be explored. Further research is needed to determine the performance of the model with the Sinhala language. However, transformers usually tend to be more data hungry than other deep learning models~\cite{marcus2018deep}. The implications of this on resource poor Sinhala may be an interesting study.

\bibliographystyle{IEEEtranN}
\bibliography{bibliography.bib}
\end{document}